# Historical Document Processing:

# A Survey of Techniques, Tools, and Trends

**James P. Philips[1*], Nasseh Tabrizi[1]**

1 East Carolina University, United States of America

*Corresponding author: James P. Philips philipsj16@students.ecu.edu

**Abstract**

Historical Document Processing is the process of digitizing written material from the past for future use by historians and other scholars. It incorporates algorithms and software tools from various subfields of computer science, including computer vision, document analysis and recognition, natural language processing, and machine learning, to convert images of ancient manuscripts, letters, diaries, and early printed texts automatically into a digital format usable in data mining and information retrieval systems. Within the past twenty years, as libraries, museums, and other cultural heritage institutions have scanned an increasing volume of their historical document archives, the need to transcribe the full text from these collections has become acute. Since Historical Document Processing encompasses multiple sub-domains of computer science, knowledge relevant to its purpose is scattered across numerous journals and conference proceedings. This paper surveys the major phases of, standard algorithms, tools, and datasets in the field of Historical Document Processing, discusses the results of a literature review, and finally suggests directions for further research.

## INTRODUCTION

Historical Document Processing is the process of digitizing written and printed material from the past for future use by historians. Digitizing historical documents preserves them by ensuring a digital version will persist even if the original document is destroyed or damaged. Moreover, since an extensive number of historical documents reside in libraries and other archives, access to them is often hindered. Historians and scholars who desire to study rare documents must travel to the archive that possesses the original, find a printed transcription or photographic facsimile, or perhaps use a microfilmed version, if one exists. Digitization of these historical documents thus expands scholars' access to archival collections as the images are published online and even allows them to engage these texts in new ways through digital interfaces (Chandna et al 2016; Tabrizi 2008). Within the past twenty years libraries, museums, and other cultural heritage institutions have scanned an increasing volume of their historical document archives. This has intensified the need to transcribe the full-text of these archival documents. Historical Document Processing incorporates algorithms and software tools from various subfields of computer science to convert images of ancient manuscripts, letters, diaries, government records, and early printed texts into a digital format usable in data mining and information retrieval systems. Drawing on techniques and tools from areas including computer vision, document analysis and recognition, natural language processing, and machine learning, Historical Document Processing is a hybrid field. Since it incorporates multiple sub-domains of computer science, knowledge relevant to its purpose is scattered across numerous disparate journals and conference proceedings. Although summaries of research in certain subfields exist, no one has yet synthesized the history of research in this field. Moreover, those existing studies do not always address the unique challenges confronting archivists, researchers, and software developers working with vintage documents. These challenges include problems with the quality of the original documents, variable quality of the digital images, and the relative scarcity of annotated training and testing data for machine learning tools compared to the vast quantity of unlabeled data. Furthermore, the historical nature of the documents themselves contributes additional issues compared to modern documents, such as complex document layouts, rare vocabulary, archaic scripts in handwritten documents, and nonstandard typefaces, ligatures, and font degradation in printed documents (Springmann et al 2014; Christy et al 2017). An extended overview of the field of Historical Document Processing is needed as an aid to researchers seeking to advance its techniques and improve its tools and for practitioners, computer scientists and archivists, designing and implementing systems to process archaic documents. To meet this need, this paper surveys the major phases of Historical Document Processing, discussing techniques, tools, and trends. After an explanation of the authors' research methodology and the scope of this review, standard algorithms, tools, and datasets are discussed, and the paper finally concludes with suggestions for further research.

## I RESEARCH RATIONALE AND ARTICLE SELECTION CRITERIA

### 1.1 Research Rationale
This article examines the evolution of the techniques, tools, and trends within the historical document processing field over the past twenty years (1998-2018) with an emphasis on the last decade. The authors believe this extended scope is warranted: No prior study was found that comprehensively summarized the steps of Historical Document Processing for both handwritten archival documents and printed texts. Many articles have focused on one dimension of the problem, such as layout analysis, image binarization, or actual transcription. However, for practical constraints most of these articles would explicitly state the assumption that earlier phases in the digitization process had already occurred (Fischer et al. 2009). Very few discussed a full historical document processing workflow. Moreover, much of the research on different algorithms and tools is dispersed throughout a myriad of publications in different subfields of computer science. New researchers and cultural heritage archivists need a synthesis of existing work. The researchers need a synthesis to contextualize their research as they direct their efforts to the frontiers of the field or its overlooked corners. Archivists pursuing their own digitization projects need to select the appropriate tools with a full understanding of the field.

### 1.2 Article Selection Criteria
The articles examined in this study were primarily drawn from three prominent databases of computer science literature: ACM digital library, IEEE Digital Library, and Science Direct. To retrieve candidate articles, the authors used several phrase and Boolean search queries. These include "historical document processing", "historic document images", and '"historic" and "document" and "images"'. This specific search vocabulary was chosen to restrict the results to articles and conference proceedings that dealt in some way with the conversion of images of historical documents to digital text. As these articles were reviewed, additional articles were discovered through their references. Articles were selected if they discussed the relevance of their research for historical documents, proposed a methodology or tool specifically for historical documents, or applied a technique to an historical dataset. Furthermore, this research focuses on historical documents written in western languages, including classical Latin, medieval and early modern European languages, and English. This emphasis reflects the current state of historical document processing field: most of the work on historical archival documents has focused on western scripts and manuscripts. While a complete discussion of developments in non-European scripts and archival manuscripts is beyond the scope of this survey, a few comments are warranted. Some research has been done on the recognition of ambiguous Japanese kana and characters in Syriac script (Nguyen et al 2017; Dalton and Howe 2011; Howe, Dai, Penn 2017). Moreover, while the field of handwriting recognition of Indian scripts and regional languages is still nascent, Pal, Jayadevan, and Sharma produced a survey of existing techniques for modern applications and Sastry and Krishnan have done work on palm leaf manuscripts written in the Indian regional language Telugu (Pal, Jayadevan, Sharma 2012; Sastry and Krishnan 2012). From the initial collection of 300+ articles, 63 were selected for more extended use within this paper.

**1.3 Intended Audience**
Due to the authors' own background and this review's emphasis on the computer science dimension of Historical Document Processing, especially algorithms, software tools, and research datasets, the authors have envisioned other computer scientists and software developers interested in historical document preservation and cultural heritage as their primary audience. However, digital humanists and archival practitioners seeking a survey of existing research and tools for use in their own archival projects are an important audience for this survey as well. Therefore, the authors have endeavored to incorporate both an algorithmic focus intended for their fellow computer scientists and discussion of trends, tools, and datasets useful to cultural heritage practitioners and digital humanists.

After a conceptual overview of Historical Document Processing in Section 2, the quantitative results of this study that included over 300 articles and reflections on current research are discussed in Section 3.

**II HISTORICAL DOCUMENT PROCESSING TECHNIQUES AND TOOLS**
Historical Document Processing incorporates several stages or phases as the pages of a manuscript or early printed book are digitized. These phases are shown in Figure 1.

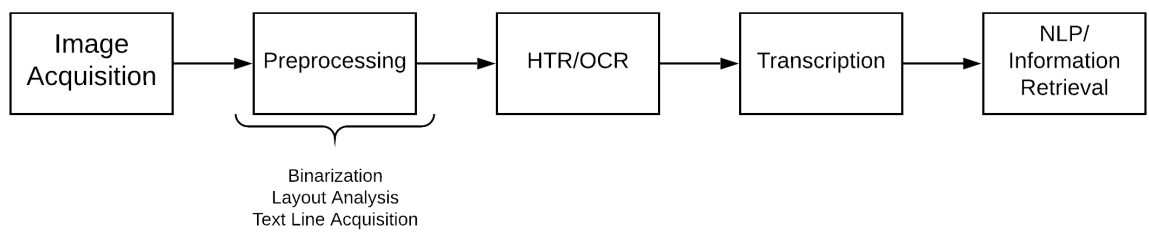

Figure 1. The steps in a conventional Historical Document Processing workflow for both handwritten and printed documents.

Once the document has been imaged, the page scans are usually preprocessed as follows:
- Binarization/grayscale thresholding
- Layout analysis/segmentation
- Text-line normalization

Following this preprocessing phase, either optical character recognition (OCR) or handwritten text recognition (HTR) is performed, depending on the type of document, using machine learning recognition software. As a data-driven step, this transcription phase, regardless of whether it is attempting a verbatim transcription of the entire document or performing isolated keyword spotting, derives the quality of its results from a combination of the quality of its training data and its underlying model.

In the subsequent parts of this section, the algorithmic techniques used for image preprocessing and textual processing are discussed. Then, software tools for ground-truth creation, text-line extraction, optical character recognition and handwriting recognition, and important datasets are examined. Finally, emergent trends in the field, such as the ascendancy of deep learning algorithms and architectures and recent historical document digitization projects, are noted.

## 2.1 Archival Document Types and Digitization Challenges

Historical documents broadly defined include any handwritten or mechanically produced document from the human past. As artifacts of previous eras, these documents furnish evidence for interpreting the textual heritage of humanity. Many have been preserved in the archives of museums and libraries, which have pursued extensive digitization efforts to preserve these invaluable cultural heritage artifacts. While early preservation efforts involved imaging, first to microfilm and then to digital (as in the Early English Books Online project)[1], an enduring goal within the field of document image analysis has been achieving highly accurate tools for automatic layout analysis and transcription of historical documents (Baechler and Ingold 2010). While historians and archivists are familiar with different types of historical documents, this section briefly discusses background on archival document types with representative example images and for the benefit of computer scientists and software developers who may be collaborating on digitization projects with cultural heritage colleagues but lack domain knowledge in historical documents themselves. The section concludes with some prevalent challenges encountered in the digitization process.

Prior to the fifteenth century and the revolution in printing technology caused by Johannes Gutenberg's movable type printing press, the majority of historical documents were manuscripts, texts composed, copied and produced by hand. After the advent and wide-spread adoption of Gutenberg's technology throughout Europe, literary works intended for publication were produced on the printing presses while private correspondence, record keeping, and other activities continued to be done by hand. This dichotomy in document types beginning in the Early Modern era and continuing to the present has led to diverse document types that must be dealt with differently during the digitization process. For example, a medieval manuscript may have a more complex layout than an eighteenth century letter, but the cursive script of the letter may be more difficult to transcribe accurately than the minuscule script of the medieval document. Manuscripts produced prior to the printing press era were written primarily on papyrus or vellum (parchment made from animal skins). While a plethora of fragmentary papyri survive from antiquity (Fig. 2), vellum supplanted papyrus as the medium of choice for documents during the medieval period. Medieval manuscripts were frequently produced in monastery scriptoria and the pages bound to produce codices. Within the cultural heritage community, significant research effort has been concentrated on producing algorithms, datasets annotated with layout ground truth, and software tools for medieval documents while in contrast less effort has been focused on ancient papyri or early modern handwritten documents (cf. Section 3).

---

[1] For background on the history and progress of the Early English Books Online project (EEBO), see https://eebo.chadwyck.com/about/about.htm#micro, Meyer and Eccles 2016, and Christy et al 2017.

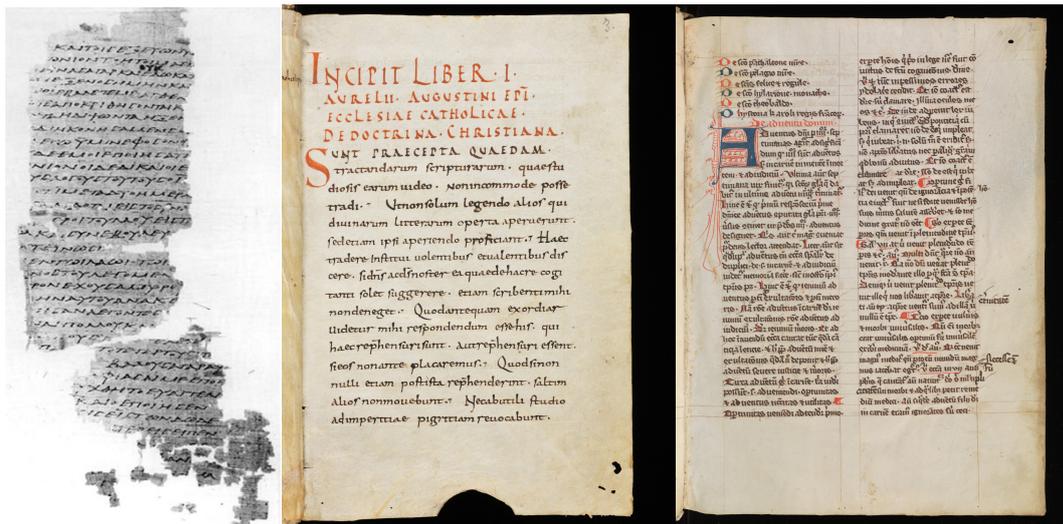

Figure 2. Fragmentary Greek Biblical Papyri from the Chester Beatty Collection (left[2]) and Medieval Latin manuscripts from the e-codices project (center[3] and right[4]) illustrating complex document layouts

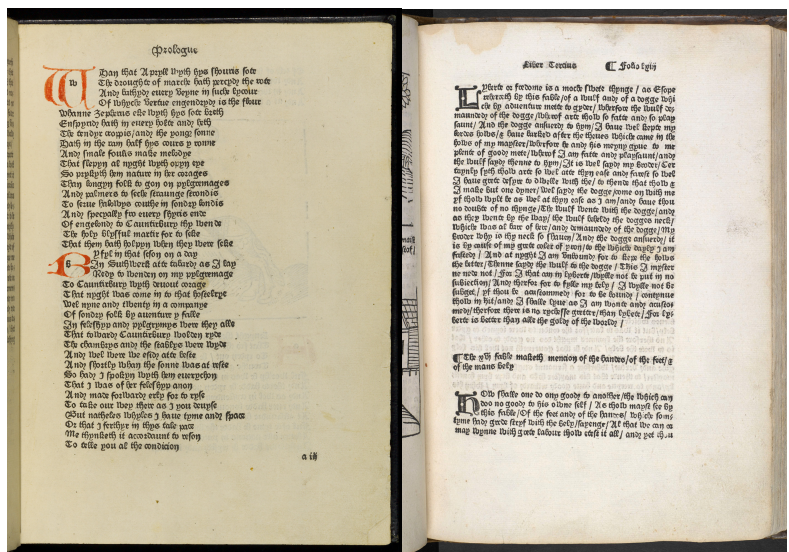

Figure 3. Late fifteenth century incunabula (left[5] and right[6]) designed to replicate medieval manuscript scripts, layouts, and ornamentation in print.

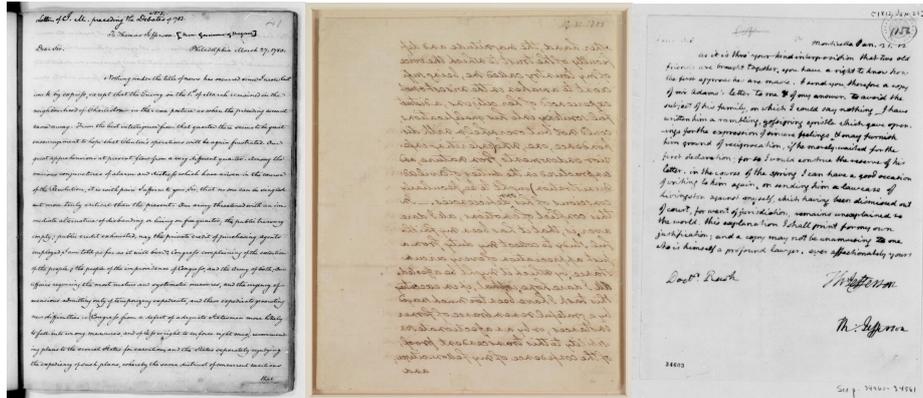

Figure 4. 18<sup>th</sup> and early 19<sup>th</sup> century letters (left and right) and speech draft (center) from United States illustrating handwriting differences and challenges in archival documents.[7]

The eclectic nature of all handwritten documents presents challenges to automatic software tools. Figures 2-4 show examples of ancient papyrus, medieval, and Early Modern handwritten documents. While organizations such as the Chester Beatty Library in Dublin, Ireland,  and the Center for the Study of New Testament Manuscripts in Plano, Texas, USA, among others have endeavored to produce digital images of archival papyri, few if any studies have been published on automatic transcription algorithms, software tools, or benchmark datasets for these ancient documents that precede the medieval period. Highly fragmentary and deteriorated, papyri will likely continue to defy automatic transcription for the near future. In contrast, codices from late antiquity and the medieval eras written on vellum tend to be much better preserved than papyri. Furthermore, their scripts are often more legible and the inter-character segmentation of minuscule are easier to train machine learning-based classifiers than the continuous cursive of early modern and modern handwritten texts. One of the primary challenges with medieval documents, however, is their complex layout (Simistira et al 2016). In addition to the primary text, these documents often feature extensive marginal or interlinear notes and textual emendations that must be identified and segmented. Likewise, these manuscripts feature lavish illustrations and ornamentation such as dropped capitals and religious artwork. In order for the primary text to be correctly transcribed, these features of medieval historical documents must be dealt with during the preprocessing phase of the HDP cycle.

Handwritten documents during the Early Modern period were frequently intended for private use or limited circulation, such as personal correspondence, record books, or diaries. The evolution of handwritten scripts and the individuality of handwriting (demonstrated in Figure 4) even for individuals whose lives overlapped the same historical era, demonstrate the complexity of HDP for these post-medieval documents. Due to the continuous cursive script, these documents are challenging during the HTR phase, while medieval documents present greater challenges during layout analysis. Like medieval documents, the primary text of these Early Modern documents must still be isolated from any marginal or interlinear annotations or illegible sections in order to produce an accurate transcription.

Regardless of the era in which they were produced, historical documents also present other challenges to the HDP process, including bleed-through from the opposite side of the page,

---

ambiguous or illegible handwriting due to the writer or document deterioration, and (depending on the circumstances in which the document was imaged) protective plastic sheets that can cause reflectivity and glare in the document image. These problems can hinder the digitization process. Moreover, some documents such as those from the EEBO were originally imaged as microfilm which was subsequently digitally imaged (Christy et al 2017); this imaging of the documents second-hand reduces the quality of the digital images that are input to the software tools.

The earliest printed texts, known as incunabula, have posed the most difficulties for accurate, digital transcription of printed works (Rydberg-Cox 2009; Springmann and Lüdeling 2017). Since these books are typeset in fonts that differ vastly from modern typefaces, modern OCR software produces poor recognition results. The extensive use of textual ligatures also poses difficulties since they declined in use as printing standardized. After 1500 greater uniformity came to printed books, and by the early 19th century, the mass production of printed texts lead to books that modern layout analysis and OCR tools could reliably and consistently digitize at scale, as seen in the digitation efforts of the Internet Archive and Google Books both in partnership with libraries (Bamman and Smith 2012).

## 2.2 Techniques

Despite differences between historical manuscripts and printed works, optical character recognition and handwriting recognition are fundamentally solutions to the same problem of text extraction. Following the preprocessing phase, text recognition either identifies keywords or creates a verbatim transcription from a line of text. Regardless of whether the text was written or printed, the central objective is to convert the words in the document image into digital text accurately. Optical character recognition relies upon the predictable regularity of space between characters and words as the basis for its classification. The character is the fundamental unit of recognition. Since the words and their constituent characters can be predictably and accurately segmented, optical character recognition classifiers can recognize and produce a transcription using the individual character glyphs. On the other hand, handwriting recognition (sometimes called handwritten text recognition) cannot rely on regular spacing of characters and words due to the idiosyncrasies of human handwriting. According to Sayre's paradox, individual letters cannot be recognized without segmentation, but the act of segmentation entails prior recognition (Fischer et al 2012). Therefore, handwriting recognition (in contrast to optical character recognition) usually relies on a recognition methodology that is segmentation free at the character level. To improve the accuracy of both optical character recognition and handwritten text recognition, systems will often incorporate a statistical language model if the document language is known beforehand (Frinken et al. 2013). A language model helps to ensure that the words recognized by the software correspond to known vocabulary and even grammar in the language. Techniques for handwriting recognition and optical character recognition alike can be organized into traditional machine learning approaches and deep learning-based approaches using neural networks. The remainder of this section discusses the preprocessing phase, text recognition techniques in historical document processing according to a taxonomy of machine learning methods, deep learning methods, and finally the related area of transcription alignment.

### 2.2.1 Preprocessing Phase

During a typical Historical Document Processing workflow shown in Figure 1, the preprocessing phase follows image acquisition and precedes any attempts to transcribe the textual content of the document. This preprocessing phase normally includes any binarization/thresholding applied to the document image, any adjustment for skew, layout analysis and text-line segmentation. Various studies have proposed various binarization methods including Bolan et al 2010, Messaoud et al 2012, Roe and Mello 2013, and Yang et al 2015. Dewarping and skew reduction methods have been proposed in studies including Bukhari et al 2011 and performance analysis conducted in Rahnemoonfar and Plale 2013. As discussed in section 2.1 above, layout analysis is one of the most challenging aspects of historical document processing. It is an especially acute challenge for medieval documents due to their complex page layouts, and many of the studies in the literature have focused on layout analysis tools, algorithms, and benchmark datasets especially for medieval documents. As Baechler and Ingold noted, earlier studies in the 2000s proposed semi-automatic tools and methodologies for layout analysis. These early tools required user interaction to identify or "annotate" document regions (Mas et al 2008). Pixel-based or connected-component-based elements in the image were utilized by these tools to identify regions in the document (Bourgeois and Emptora 2007; Ramel et al 2007; Clausner et al 2011). In their study Baechler and Ingold proposed a layout model for medieval documents that they envisioned would support "fully automatic annotation and transcription tools (275)." Using manuscript images from the e-codices project (images from this project would also be incorporated into the Diva-HisDB set discussed below) at the University of Fribourg[8], they modeled a document page in a medieval manuscript based on several "layers", including document text, marginal comments, degradation, and decoration. Overlapping polygonal boxes are used to identify the constituent layers and are represented in software via XML.

In their work, Gatos et al 2014 developed a layout analysis and line segmentation software module designed to produce input to Handwritten Text Recognition (HTR) tools. Their work was incorporated into the Transcriptorium project's Transkribus software (cf. Sec. 2.3.5). Their tool first identifies "text zones" using a combination of any vertical ruled lines in the document, vertical sections of white space ("white runs"), and finally horizontal restriction/refinement. This enables their method to detect document sections separated by lines, multi-column document layouts separated by runs of white space, and the use of horizontal lines to achieve more precise identification of the text zones. The text lines are then identified in each textual region by calculating the average character height, applying a Hough transform mapping that "[detects] lines that intersect with the connected components of each line (Gatos et al 2014, 467)." Using sample document images from the Transcriptorium project, they created a benchmark dataset to evaluate the performance of their layout analysis module and its methodology, achieving an 84.7% accuracy rate of identifying primary text zones. They also evaluated the accuracy of their text-line segmentation approach using document images from the Transcriptorium project. On these they achieved an 83.08% detection rate, 86.35% recognition accuracy, and an F-measure of 84.68%, outperforming previously proposed methodologies (Gatos et al 2014, 469).

In a pair of detailed studies, Pintus, Rushmeier, and Yang likewise explore layout analysis and text-line extraction with an emphasis on medieval manuscripts. Echoing the need for automatic layout analysis cited by Baechler and Ingold and Gatos et al, Pintus et al 2015 in their first study address the problem of initial calculation of text-line height. They note the process of estimating this essential metric is exacerbated by the overlap of descenders and ascenders from adjacent text-lines with narrow inter-line spacing encountered often in medieval manuscripts. They proposed an automatic text-line extraction algorithm that computes the line-height per page of a manuscript. Their method computes text-height by creating a multi-scale representation of the document image that produces a collection of sub-images. Each of these is individually processed using NACF, projection profiles, discrete Fourier transform, and the results of this PMF are summed to compute the estimate of line height. Using this value, they then segment the text regions coarsely and apply a SVM classifier to produce a refined text line identification. They note their method is not adversely affected by skewed texts and usually does not necessitate any alignment correction. On test datasets of 15,552 pages and 80,000 text lines, they achieved 98.55% average precision and 96.31% average recall for their text line segmentation. Moreover, they provided time metrics for their algorithms' runtime. Using multi-threaded execution, each page required an average of 3 minutes processing time.

---

[8] https://www.e-codices.unifr.ch/en

In their second study, Yang et al (2017) extend their work on text-height estimation and layout analysis to an automated system that can work on a per-page basis rather than per manuscript (a limitation they cite regarding their previous work). They propose three algorithms, one for text-line extraction, one for text block extraction, and one for identifying "special components." These use semi-supervised machine learning technique and focus on medieval manuscripts produced originally by professional scribes. Thus, Yang et al note their algorithms are not designed to deal with warped pages or severely damaged documents (4). For text block computation, they achieve 97% precision and 96% recall on their test set of 35 images containing 42 text-blocks. On an extended dataset experiment, they reported 99% precision with 96% recall for 2,045 text blocks. One drawback to their work is that they do not evaluate their algorithms against the Simistira et al dataset or the results of Gatos et al. Nevertheless, their results are impressive and demonstrate that the desideratum of automatic algorithmically-layout analysis with high precision, recall, and accuracy is drawing nearer to reality.

### 2.2.2 Handwritten Text Recognition Techniques

Although historical handwriting recognition has been extensively researched, data-driven techniques using both traditional machine learning and deep learning have dominated recent research. Given the inherent challenges of handwritten text recognition, especially for historical documents, some studies including (Rath and Manmatha 2006; Fischer et al. 2012) explored keyword spotting techniques as an alternative to the production of a complete transcription. Early keyword spotting techniques applied to historical documents as an image similarity problem. In this method, clusters of word images that have been compared for similarity using pairwise distance are created. Those clusters that correspond to significant words in the document are then manually labeled with their word. The labels can then be indexed and queried in an information retrieval system. This approach to keyword spotting is also sometimes known as template-based matching. In their study Rath and Manmatha used a dynamic time warping algorithm to compute image similarity and compared the performance of several clustering algorithms including Ward linkage and k-means on an early version of the George Washington dataset. However, their best word error rate was 38.12%. As part of the HisDoc project, Fischer et al explored several data-driven techniques for both keyword spotting and complete transcription (Fischer et al. 2009, 2012, 2014). One problem with word-based template matching is that the system can only recognize a word for which it has a reference image. Rare (out of vocabulary) words cannot be recognized. As a solution to this limitation, the HisDoc researchers applied character-based recognition with Hidden Markov Models to keyword spotting. They noted: "When the learning-based approach is applied at the character level, a word spotting system obtains, in principle, the capacity to spot arbitrary keywords by concatenating the character models appropriately (Fischer et al 2012)." For their keyword spotting analysis, they compared the character-based system with a baseline dynamic time warping system. Using mean average precision as their evaluation metric, they found that the Hidden Markov Models significantly outperformed the Dynamic Time Warping system on both localized and global thresholds for the George Washington and Parzival datasets (GW: 79.28/62.08 vs 54.08/43.95 and Parzival 88.15/85.53 vs 36.85/39.22). The HisDoc project also compared Hidden Markov Models and neural network performance on the three datasets comprising the IAM-HisDB (this includes the St. Gall, Parzival, and George Washington datasets) to produce full transcriptions. The Hidden Markov Model-based system was identical to that used in their earlier keyword spotting study. For their recurrent neural network-based system, they used a bidirectional long short-term memory architecture that could mitigate the vanishing gradient problem of other neural network designs. Each of the nine geometric features used for training corresponds to an individual node in the input layer of the network. Output nodes in the network correspond to the individual characters in the character set. The probability of a word is computed based on the character-probabilities. According to (Fischer, Naji 2014), word error rates were significantly better for the neural network architecture than the Hidden Markov-based system on all three sets of historical document images: St. Gall 6.2% vs 10.6%, Parzival 6.7% vs 15.5%, and George Washington 18.1% vs 24.1%.

Neural networks continue to be the ascendant technique within the field for historical handwritten text recognition. For example, (Granell et al. 2018) examined the use of convolutional recurrent neural networks for late medieval documents. As they note in their study, the convolutional layers perform automatic feature extraction which precludes the need for handcrafted geometric or graph-based features such as those used by HisDoc. The trade-off is that for deep neural network architectures to be competitive for time efficiency with other techniques, they require significant computational power. Usually this is obtained through the use of a graphical processing unit (GPU) rather than a CPU. Working with the Rodrigo dataset, they achieved their best results using a convolutional neural network supplemented with a 10-gram character language-model. Their word error rate was 14%. The *In Codice Ratio* project has likewise used a neural network in their efforts to transcribe papal correspondence in the Vatican archives.

### 2.2.3 Historical Optical Recognition Techniques

As with handwriting recognition, historical optical character recognition can be accomplished with several techniques. However, neural network-based methods have become more prominent in the software libraries and literature recently. Since printed texts in western languages rarely use scripts with interconnected letters, segmentation-based approaches are feasible with optical character recognition that are not practical for handwritten text recognition. Nevertheless, historical optical character recognition is drastically more difficult than modern optical character recognition (Springmann and Lüdeling 2017). One challenge is the vast variability of early typography. Not only were historical printings not laid out with modern, digital precision, but also a plethora of early fonts were utilized since each printer usually created his own sets of type. This led to a proliferation of font designs across the British Isles and continental Europe (Christy et al 2017). This means that a multitude of typeface families exist, including Gothic script, Antiqua, and Fraktur (for Germanic texts). Although more mechanized printing techniques were developed in the early nineteenth century, early modern printing from the mid-fifteenth century through the eighteenth century is too idiosyncratic for optical character recognition systems trained using modern, digital fonts. Among the most difficult historical texts for optical character recognition are incunabula due to their extensive use of ligatures, typographical abbreviations derived from medieval manuscripts that do not always have a corresponding equivalent in Unicode, and unpredictable word-hyphenation across lines (Rydberg-Cox 2009). A few approaches have been proposed to circumvent the challenge of historical typography. The training limitations of commercial software such as Abbey Fine Reader mean that researchers must resort to open source alternatives such as Tesseract or OCRopus (Springmann et al. 2014). Tesseract's classifier can be trained using either synthetic data (digital fonts that resemble historical ones) or with images of character glyphs cropped from actual historical text images. However, Tesseract does not include any built-in preprocessing tools. Furthermore, as originally developed, Tesseract did not implement a neural network-based classifier. An update to OCRopus, however, was designed with a recurrent neural network architecture (Breuel et al. 2013). Like the HisDoc classifier, OCRopus uses the bidirectional long-term short-term memory network. Although high accuracies are achievable with this architecture, some of the same caveats apply from its use for handwritten text recognition. The classifier requires substantial training data, with the corollary of extensive ground truth that must be created manually, and this classifier is computationally intensive for CPUs (Springmann et al 2014). Some of the most thorough and clearly presented work on historical optical character recognition using OCRopus was done by Springman and Lüdeling on an eclectic, diachronic corpus of printed works ranging from 1487 to 1870 (Springmann and Lüdeling 2017). They examined training separate models for each book in the corpus as well as mixed models that included representative texts across the chronological continuum of their corpus in the training set. They consistently achieved character accuracies between 96.3% and 99.6% for each of the 20 books in the collection using the trained model corresponding to each book. The mixed models had mean character accuracies of 95.81% on one subset of their corpus and 94.27% on the other. In evaluating the accuracy of the optical character recognition results, they note that the threshold of acceptable accuracy will vary based on the purpose of the digitization. While a 95% accuracy rate would be sufficient for some information retrieval tasks for which recall is to be prioritized over precision, achieving higher accuracies for more precise use cases may require post-correction of the output. Furthermore, it may require solutions to problems inherent in historical texts apart from text recognition, such as the evolution of word forms from the text's historical era to the present and the lack of standardized spelling in historical documents. Nevertheless, their use of a tool that implements a recurrent neural network-based technique to historical optical character recognition of a diachronic corpus is a significant achievement. Not only was high accuracy achieved, but their work demonstrates the ascendancy of neural network-based architectures for both kinds of historical document processing: optical character recognition and handwriting recognition.

## 2.3 Tools

Several software tools and datasets exist for researchers and practitioners pursuing historical document processing. For historical text optical character recognition, these include the Abbey FineReader, Tesseract, OCRopus, and AnyOCR tools and primarily the IMPACT dataset of early modern European printed texts. Few generic tools exist for historical handwriting recognition tasks, but researchers do have access to the IAM-HistDB, Rodrigo, and In Codice Ratio datasets. These variously contain images of full manuscript pages, individual words and characters, and corresponding ground truth for medieval Latin and early German and Spanish manuscripts. The IAM-HistDB also contains the Washington dataset for historical cursive handwriting recognition. In addition to software and datasets for the transcription phase of historical document processing, the Alethia tool and the IMPACT and Diva-HistDB datasets can be used for researching layout analysis and other preprocessing tasks. The rest of this section surveys the characteristics of the available software tools and datasets and concludes with a discussion of training, testing, and evaluation methodologies that form a common foundation for these data-driven tools.

### 2.3.1 Software Tools

As discussed in Section 2.1, the eclectic nature of historical typography and printing, especially the eccentric variety of early modern typefaces, thwarts modern optical character recognition software. Tools such as the commercial Abbey Fine Reader and open source Tesseract were originally developed for modern documents. However, some researchers and archival projects have repurposed both tools for use in historical document processing. For example, many of the digitized literary works on the Internet Archive were transcribed with Abbey Fine Reader, and Google's Books initiative has utilized Tesseract. More recently, the European IMPACT project (cf. Section 3.3) used Abbey Fine Reader in conjunction with the IBM Adaptive OCR Engine, and Ul-Hasan et al used Tesseract as part of their toolchain to generate training data for another open source tool, OCRopus (Christy et al. 2017; Ul-Hasan et al. 2016). The eMop project (cf. Section 3.3) likewise used Tesseract for transcribing microfilmed images of seventeenth century printed texts.

Development of Tesseract[9] began in the mid 1980s (Smith 2006). After a hiatus in development, the source code was released as open source in 2005. It has become one of the established tools in the field, and a strong rival to commercial tools such as Abbey Fine Reader due to its input flexibility (Christy 2017). During operation the tool first identifies text lines, even skewed or slanted lines, fits the baseline, performs proportional word recognition, and separates joined characters (Smith 2006). After these steps are complete, a static classifier creates a list of potential character matches, computing the bit vector similarity between the unknown character and each potential match. Tesseract conducts two passes over the data during its word recognition phase. The words recognized with a high confidence rate in this initial pass become training data for an adaptive classifier that then re-examines and classifies the unrecognized words. Tesseract thus combines semi-supervised machine learning with a segmentation-based approach. Ray Smith, the tool's creator, notes that it uses minimal linguistic analysis in forming its decision for each word's classification. Written in C/C++, Tesseract has remained in active development since its source code was first released. The tool is entirely command-line based and includes a API that permits its integration into other workflows. Subsequent updates have incorporated LSTM neural networks. Researchers seeking to integrate Tesseract with the Python programming language have the PyOCR library available.[10]

OCRopus, OCRoRact, and anyOCR are newer tools for optical character recognition. Originally developed at the German Research Center for Artificial Intelligence at the University of Kaiserslautern, all three software projects are open source and built on extensible, modular design. Like Tesseract, OCRopus was first created for text recognition in modern documents and later applied to historical ones (Springman et al 2014). Following its preprocessing phase, which the software supports through modules for binarization, noise removal, skew correction, page region segmentation, and layout analysis to detect columnar layouts, OCRopus performs textline recognition (Shafait 2009). Individual characters are the basis for segmentation and subsequent classification. The individual characters are classified or recognized using a classification module that implements a finite state transducer and represents hypothetical segmentations as graphs. The classifier bases its decision on the traversal of the graph with the lowest cost. Through the use of this classification method, OCRopus is able to classify more accurately anomalous instances due to scanning artifacts and better distinguish upper and lower-case characters. Language model integration is supported, which can further enhance the character classification results. However, the absence of language models for historical languages continues to be a problem (Springmann 2014). The modular design of OCRopus permits language models to be easily swapped into the software's workflow. After its initial creation, OCRopus was renamed OCRopy.[11]

---

[9] https://github.com/tesseract-ocr/tesseract
[10] https://gitlab.gnome.org/World/OpenPaperwork/pyocr

To improve transcription accuracy, OCRoRact incorporates both Tesseract and OCRopus into a single system. One motivation for its creation was to circumvent the need for extensive ground truth data, which language experts must often create manually through an expensive, time-intensive process (Ul-Hassan 2016). Minimal ground truth is used to train Tesseract. These results are then used to train the OCRopus classifier. Several iterations of training follow until the rate of improvement between iterations is less than 1%. Through the combination of Tesseract, which uses a segmentation-based classifier, and OCRopus, which uses a segmentation-free classifier, Ul-Hasan et al were able to achieve a character error rate (CER) on par with OCRopus while alleviating the need for extensive, manually annotated ground truth.

In their work on anyOCR, Jenckel and Bukhari et al extended the hybrid approach taken with OCRoRact (Jenckel et al 2016; Bukhari et al 2017). Motivated by the same goal of minimizing or even eliminating the need for manually-generated ground truth, the creators of anyOCR substituted their own unsupervised, segmentation-based classifier in place of Tesseract. Since an unsupervised classifier does not require any ground truth annotation, this approach is ideal for historical document processing since the features are algorithmically determined from the training images rather than explicitly specified. The classifier uses the researchers' own variant of the k-means clustering algorithm that iteratively clusters the character images based on a "blurriness" metric. While the reported character error rate in their experiments on a fifteenth century printed Latin text was only marginally better than the results achieved by OCRopus on the same dataset, the use of an unsupervised classifier for the segmentation-based phase significantly reduced the time contribution needed from language experts.

### 2.3.2 Datasets

Despite the extensive scale of the historical document processing task, datasets for training, testing, and evaluation remain scarce in comparison to those used in related fields such as modern handwriting recognition. For Western historical documents, research datasets exist for medieval Latin, medieval German and Spanish, a variety of early modern European languages, and eighteenth-century English. This section describes the datasets for historical handwriting recognition, optical character recognition, manuscript and printed document layout analysis, and finally the published data of the *In Codice Ratio* project. A hierarchy of HDP datasets is shown in Figure 5.

---

[11] https://github.com/tmbdev/ocropy

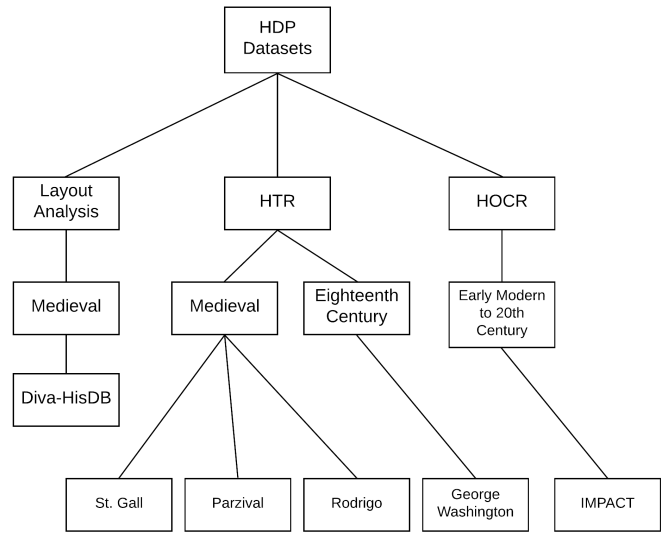

Figure 5. A Hierarchy of HDP Datasets Based on use case and time-period

Few options exist for researchers seeking to work with medieval manuscript transcription. Two medieval datasets are included in the IAM Historical Document Database (IAM-HistDB).[12] One of these, the St. Gall dataset, features images of a ninth century Latin manuscript written in Carolingian script by a single scribe.[13] The original manuscript, a hagiography of St. Gall, is from the Abbey of St. Gall in Switzerland. Fischer et al utilized the images and corresponding page transcriptions from J.P. Migne's *Patrologia Latina* previously published to create the dataset (Fischer et al. 2011). In addition to the page images and transcription, the dataset includes extensive ground-truth: text-lines and individual word images have been binarized, normalized, and annotated with verbatim line-level transcription. Originally developed by the HisDoc project (cf. 2.3.2) for its work on transcription alignment with archival document images, the dataset has since been used in further research (cf. 3.7).

While Latin was the dominant ecclesiastical and scholarly language of Europe during the medieval period, some literature was produced in the vernacular languages. Two datasets exist for researchers investigating handwriting recognition in those vernacular manuscripts, specifically the Old German and Old Spanish or Castilian dialects. Also part of the IAM-HistDB collection, the Parzival dataset contains manuscript pages of an Arthurian epic poem written in Old German from the thirteenth and fifteenth centuries.[14] In contrast to the St.Gall manuscript which was produced by a single scribe, the forty-seven Parzival images are drawn from three different manuscripts produced by three scribes using Gothic minuscule script in multi-column layouts. Like the St. Gall set, the Parzival collection includes page images and transcription along with ground truth annotation. Text-lines and single word images have been binarized, normalized, and annotated with a full line-level transcription. Known as the Rodrigo corpus[15], the Old Spanish dataset is longer than either the St. Gall or Parzival datasets at 853 pages. Created like the preceding datasets for handwriting recognition and line extraction research, the researchers based at the Universitat Politecnica de Valencia used the digitized images of an Old Spanish historical chronicle, the "Historia de Espanana el archbispo Don Rodrigo (Serrano et al 2011)." Although the work is fourteenth century, the manuscript in the dataset is from 1545, and thus can be traced to the threshold of the emergence of printing press technology. Although the creators of the dataset published results of running a hybrid Hidden-Markov model-based image classifier with a language model, Granell et al have used the dataset with deep neural networks (Granell et al 2018).

The Washington dataset is the third dataset included in the IAM HistDB.[16] The collection includes a selection of correspondence written by George Washington and his secretary. The original letters were drawn from his papers at the United States Library of Congress. The script is continuous cursive in the English language. First used in Rath and Manmatha's research on word spotting (cf. Sec. 3.2), the HistDoc project augmented the dataset with individual word and text-line images and corresponding ground truth transcriptions for each line and word (Fischer et al 2010). The Washington dataset is unique. It is the only historical document dataset designed for cursive handwriting recognition, and therefore is a particularly valuable tool.

While the preceding datasets for medieval handwriting recognition were limited to a handful of manuscripts written in Latin and early vernacular languages from only a few countries, researchers and archivists working with early modern optical character recognition have a much more comprehensive resource in the IMPACT dataset (Papadopoulos et al 2013). Created by a consortium of European libraries, the IMPACT dataset provides a truly diverse, pan-European collection that includes texts from the fifteenth through the twentieth centuries in eighteen different European languages and a variety of distinct scripts. Printed books comprise the majority of the collection, but it also includes newspaper pages, legal documents, journals, and an assortment of miscellaneous documents. Focused on creating a dataset that represented the types of printed archival documents held by the different contributing libraries, the researchers hoped "that the IMPACT image and ground truth dataset will continue to be the basis for cutting-edge research related to digitization and OCR (Papadopoulos et al 2013, p. 123)." In addition to document metadata, the extensive set of ground truth for this collection includes both the full document text in Unicode along with layout analysis annotation and reading order specified with the XML-based PAGE format. The Unicode-based transcription of the page text permitted accurate encoding of the variety of printed scripts within the dataset and ligatures, special abbreviation characters used extensively throughout early modern texts. Despite its value for early modern optical character recognition, however, as Springmann et al have observed, the IMPACT dataset does have a limitation regarding Latin texts. Latin remained the common language of scholarly discourse into the eighteenth century, but Latin texts are only a small part of the dataset (Springmann et al 2014). Nevertheless, the IMPACT dataset is a comprehensive resource for early modern vernacular optical character recognition.

The previously described IAM-HistDB datasets dealt exclusively with historical handwriting recognition. As a benchmark for evaluating preprocessing performance on medieval documents, the HistDoc project created the Diva-HistDB.[17] This dataset contains 150-page images from three different manuscripts with accompanying ground truth for binarization, layout analysis, and line segmentation (Simistira et al 2016). Written in Carolingian script, two of the manuscripts are from the 11th century, and one from the fourteenth century written in Chancery script. All three manuscripts have a single column of text surrounded by extensive marginal annotation. Some pages have decorative initial characters. The layouts are

---

thus highly complex. The ground truth concentrates on identifying spatial and color-based features. Like the IMPACT dataset, the ground truth is encoded in the PAGE XML format.[18] The dataset is freely available on the HistDoc project website.

Finally, the *In Codice Ratio* project (cf. Section 3.3) should also be noted. The *In Codice Ratio* team have created their own dataset for training their classifier on handwritten medieval characters and testing their project (Firmani et al. 2017; Firmani et al 2018).[19] In contrast to the other medieval manuscript datasets, their dataset focuses solely on the individual characters and corresponding annotation. Since the manuscripts they were digitizing contained twenty-three distinct characters, they extracted characters from two representative manuscript pages. Following some synthetic augmentation to create additional character examples for training, the dataset contained one thousand-character examples for each character class. These sample images are included in one dataset, while ground truth consisting of word PNG images and associated text-file transcriptions are included in a separate dataset. Both are freely downloadable from their project website.

While most of the handwriting and optical character recognition datasets discussed in this section have focused on Latin languages or Latin script, a dataset has been created for HWR and OCR of historical polytonic (i.e. multiple accent) Greek texts. Introduced by Gatos et al, the dataset was developed for research on word and character recognition as well as line and word segmentation (Gatos et al 2015). It features 399 pages of both handwritten and printed Greek text, mostly from the nineteenth and twentieth century.[20]

### 2.3.3 Methodologies for Training, Testing, and Evaluation

As with any data-driven digital inquiry, historical document processing requires methodologies for verifying the accuracy and performance of the algorithms and software tools used. Data integrity is an imperative for this field because historical interpretation is dependent on the virtue of veracity. For handwriting recognition and optical character recognition alike, most studies and tools utilize either conventional machine learning techniques or neural network-based techniques. This means that annotated transcription data or "ground truth" is needed for the character or word classifier to map the text in the document image to its transcription during the training phase. Most current tools use supervised machine learning which requires the annotated data for training and evaluation. An exception is the Ocular OCR engine from the University of California-Berkley's Natural Language Processing Lab (Berg-Kirkpatrick et. al 2013). This software utilizes an unsupervised classifier. Regardless, the annotated data is still necessary for the testing phase to measure the tool's accuracy and performance on unseen data. Some studies follow a conventional tripartite partitioning of the dataset into training, testing, and development subsets. Others apply cross-fold validation techniques. Some datasets have ground truth for layout analysis as well, such as the IMPACT and Diva-HisDB datasets. OCRopus includes a tool for entering a ground truth transcription for extracted text lines on an html page. Since the scarcity of quality annotated data remains a concern, some of the researchers have proposed solutions. Fischer et al. discussed their creation of ground truth for the IAM-HisDB (2010). Their workflow included a combination of automated and manual steps. To alleviate the significant time investment to produce annotated ground truth, the In Codice Ratio researchers used a crowdsourcing approach with a Web-based interface that presented positive and negative character images and prompted the users to classify a small set of character images. Wei et al examined the use of Gabor-features to generate ground truth for document layout. Another approach has been studied by Fischer et al: algorithmic transcription alignment. Rather than having to enter the transcription manually, Fischer proposed leveraging existing transcriptions of scanned documents to produce annotated corpora more quickly. They used a Hidden Markov Model recognizer to perform alignment.

---

[18] The PAGE format (Page Analysis and Ground-Truth Elements) is described in Pletschacher and Antonacopoulos 2010
[19] The test dataset developed by In Codice Ratio is available from the project website: http://www.inf.uniroma3.it/db/icr/datasets.html
[20] GRPOLY-DB can be downloaded from: http://users.iit.demokritos.gr/~nstam/GRPOLY-DB/

Several metrics are used to evaluate the performance of an historical document processing system. For handwritten text recognition systems that use image similarity, precision and recall are two important performance measures. Precision ascertains how many of all the relevant results in the dataset were actually retrieved. For machine learning systems, transcription performance is evaluated using character error rate, word error rate, or sometimes both if a language model is utilized to enhance the recognition results. Layout analysis performance is assessed using line error rate and segmentation error rate (Bosch et al 2014).

### 2.3.4 Software Systems for Historical Document Processing

Cultural heritage practitioners seeking production-ready tools for their own historical document preservation projects or research have two software systems available that provide a full suite of tools for preprocessing, machine learning training, and transcription. These two tools are DIVA-Services (Würsch et al 2017) and the Transkribus platform from the EU-sponsored READ project (Kahle et al 2017).

Developed by researchers at the University of Fribourg, DIVA-Services is a web-based service built on a RESTful architecture that provides a JSON-based API (application programable interface) of tools for each stage of historical document processing. Würsch et al note that providing common document image analysis (DIA) and machine learning algorithms and text recognition tools via an API or web-based interface decreases the burden on computer science researchers and cultural heritage practitioners who wish to use existing tools without the complicated steps of configuring a plethora of different and potentially mutually incompatible software libraries. Abstracting tools behind a consistent API enables researchers and practitioners to incorporate these tools into their own applications with less effort. Thus, DIVA-Services offers a suite of standard DIA tools to HDP developers to include in their own HDP toolchain, including:

- Difference and Laplacian of Gaussian binarization methods
- Pixel and interest point layout analysis methods for text-line and image extraction
- Local binary, scale-invariant feature transform (SIFT), and Gabor features for machine learning feature extraction
- OCRopus and Tesseract libraries for optical character recognition
- Implementations of support vector machine (SVM), Gaussian mixture methods, and K-nearest neighbors (KNN) machine learning classification algorithms.

Since DIVA-Services does not furnish a complete integrated HDP toolchain but rather a collection of HDP services for DIA, HTR, and OCR, it is better suited to research use-cases or archival projects that need to take an eclectic approach to tools in a customized software toolchain. Moreover, as an open source project, it is also an excellent choice for researchers and practitioners who hold to an open source ethos in their tools and research. Since it is fully open source, there are no costs due to commercial software licensing fees and the tool chain is completely transparent for evaluation purposes.

Originally Created by the EU's Transcriptorium project and continued by the READ project, the Transkribus software platform endeavors to provide a full HDP toolchain to computer science researchers, cultural heritage scholars, and archival practitioners. Described by Kahle et al, the Transkribus platform consists of a software client that communicates with a central server through a RESTful API. Users utilize the software client on their desktop computer to upload document image files and any pre-existing layout or transcription data that has been stored in the PAGE XML or ALTO formats. The client allows users to segment document images manually or use automatic layout analysis functions. Authorized users on the service also have access to training machine learning models for new datasets in both HTR and OCR. The Transkribus platform integrates existing layout analysis tools from the Technical University Vienna's Computer Vision Lab and Gatos et al 2014. For OCR it uses the ABBY Fine Reader SDK. Two methods are offered for HTR: a HMM-based tool from the University of Valencia's Pattern Recognition and Human Language Technology research group that incorporates a language model, and a RNN-based tool that lacks the out-of-vocabulary limitation of the HMM-LM tool. Since it uses a RESTful API, third party clients can utilize the Transkribus platform. Through the planned additions of a web-based interface to supplement the desktop client, the service will support crowd-sourcing transcription projects. Some parts of the Transkribus architecture are available as open source, but others are not such as the third party DIA tools used on the server backend.

### 3.3.5 Evaluation of DIVA-Services and Transkribus

DIVA-Serices and Transkribus offer similar feature sets to the cultural heritage community. However, they should not be seen as direct competitors. As a cross-platform software service, Transkribus is likely the better solution for archivists seeking an integrated HDP toolchain that requires minimal or no custom software to be developed. Since it offers multiple tools for each step in the HDP process and supports standard formats such as PAGE, it is ideally suited for archivists who need a reliable service for an historical document transcription project that allows support for machine learning training on new datasets. Due to the platform's hybrid open source-closed source nature and lack of tool modularity (users cannot substitute their own libraries directly for a Transkribus one), users who need more flexibility and alignment with open source values may find DIVA-Services more suited to their needs. Since DIVA-Services provides separate API calls for each discrete step in the HDP workflow, this service is more suitable for computer science researchers and archivists who need to integrate existing methods alongside custom software. DIVA-SERVICES and Transkribus thus offer complementary approaches that meet the different use cases of members of the cultural heritage community.

### III Trends in Recent Digitization Projects

Within the past decade several research projects have advanced the field of historical document processing through the creation of datasets, the exploration of improved techniques, and the application of existing tools to digital archival document preservation efforts. Each project has contributed to the field in unique and complementary ways. Earlier sections of this survey article have discussed noteworthy techniques and datasets developed by these projects. The remainder of this section summarizes the projects, the purpose of their work, and concisely reviews their contributions. These details furnish historical context to the research and inspiration for further research efforts.

The HisDoc family of projects have made significant contributions to algorithms, tools, and datasets for medieval manuscripts. The inaugural HisDoc project[21] was a collaborative effort among three Swiss universities: the University of Fribourg, the University of Bern, and the University of Neuchatel. Lasting from 2009 to mid-2013, the project concurrently studied three phases of historical document processing: layout analysis, handwriting recognition, and document indexing and retrieval (Fischer Nijay et al 2014). While much of their research focused on medieval documents and scripts, their goal was to create "generic methods for historical manuscript processing that can principally be applied to any script and language (p. 83)."

Each team concentrated on creating a module for each phase. The researchers at Fribourg focused on layout analysis.[22] They created a tool to classify the various elements of a manuscript page, including background and foreground, text and illustrations, and decorative elements such as ornamental first characters (p. 86). Describing their methodology as a "pyramidal approach", parts of the page layout were identified successively at different image resolutions with a neural-network-based classifier: foreground and background were demarcated at low resolution with a downscaled image and text and non-text elements were distinguished at higher resolution. Further refinement transformed the text elements into lines and words and non-text elements into illustrations and ornamental designs. Meanwhile, the research team at Bern concentrated on the transcription phase (p. 88).[23] They explored multiple approaches to the transformation of text-line images into digital text, including keyword spotting and full transcription, and multiple techniques such as traditional machine learning strategies (hidden Markov models and graph-based features) and neural networks (bidirectional long-term short-term memory and recurrent neural networks) (Fischer Reisen 2010). Moreover, they pursued the use of statistical language models and corpora to improve transcription accuracy. The work of the Bern team also investigated transcription alignment to pair an existing transcription with its corresponding location in a text-line (Fischer 2011). This was proposed as a potential, partial solution to the dearth of labeled data. With the aim of developing a search engine for historical manuscripts, the HisDoc team at Neuchatel investigated various challenges of information retrieval in the context of historical documents.[24] For a particular search query, their system would produce a list of matching manuscripts ranked by relevance to the original query (p. 90). Their research investigated not only the mitigation of latent word recognition errors introduced during the transcription phase but also difficulties inherent in historical texts such as inflectional and orthographical variety. To solve the transcription error problem, alternative transcription hypotheses with high probabilities were used to augment the main transcription, becoming part of the text indexed by the information retrieval system. The researchers also conducted performance evaluation of standard probabilistic, vector-space, and language-based information retrieval models on the historical document transcriptions. Information retrieval is the ultimate goal of historical document processing, and the team at Neuchatel expanded the knowledge of the field through their research. Through this concurrent exploration of layout analysis, handwriting recognition, and information retrieval, the first HisDoc project was a pioneering endeavor that advanced multiple aspects of the field. Especially important was their recognition of the need for a modular, unified system for historical document processing and the creation of the important IAM-HistDB dataset.

---

[21] https://diuf.unifr.ch/main/hisdoc/
[22] https://diuf.unifr.ch/main/hisdoc/module-1-layout-analysis
[23] https://diuf.unifr.ch/main/hisdoc/module-2-handwriting-recognition
[24] https://diuf.unifr.ch/main/hisdoc/module-3-information-retrieval

HisDoc 2.0 was conceived as a direct extension of the original HisDoc project.[25] Concentrated at the University of Fribourg, the focus of this project was advancing digital paleography for archival documents (Garz et al 2015). The HisDoc 2.0 researchers recognized that historical manuscripts are complex creations and require multi-faceted solutions from computer science. Frequently written by multiple scribes and haphazardly-arranged page layouts, many documents do not conform to the ideal characteristics explored during the first HisDoc project. With HisDoc 2.0, the researchers investigated combining text localization, script discrimination, and scribal recognition into a unified system that could be utilized on historical documents of varying genres and time periods. Additional goals of the project were the automatic generation of paleographic descriptions derived from the system output and the incorporation of existing semantic descriptions of documents into the system algorithms. Lasting from 2014 to 2016, the HisDoc 2.0 project made several contributions to the field. One was DivaServices, a web service offering historical document processing algorithms with a RESTful (representational state transfer) API to circumvent the problem many developers and practitioners face with the installation of complicated software tools, libraries, and dependencies (Würsch et al 2016). Another contribution was the DivaDesk digital workspace, GUI-based software that makes computer science algorithms for ground truth creation, layout analysis, and other common tasks accessible for humanities scholars (Eichenberger et al 2014). The project explored ground truth creation, text region and layout analysis with neural networks, and aspects of writer identification. Finally, the project produced and released the Diva-HisDB dataset (cf. Section 3). The spirit of the two HisDoc projects has continued in a third, recent initiative, HisDoc III.[26] It began in 2017 and seeks to develop deep-learning-based methods to classify the voluminous number of unlabeled manuscripts in archives and to create workflows that could be implemented by libraries and other cultural heritage archives.

The IMPACT project (Improving Access to Text)[27] was a European Union-funded initiative to develop expertise and infrastructure for libraries digitizing the textual heritage of Europe. Despite the rapid rate of text digitization by European libraries, the availability of full text transcriptions was not keeping pace. With many libraries solving the same digitization challenges, solutions to problems were being duplicated, leading to inefficient use of time and resources. Moreover, the cost of manual transcription was prohibitive (estimated at 400 to 1000 euros per book, depending on the book length).[28] Finally, existing optical-character recognition software produced unsatisfactory levels of accuracy for historical printed books. Through the formation of a pan-European consortium of libraries, the IMPACT project consolidated digitization expertise and developed tools, resources, and best practices to surmount the challenges of digitization on such an extensive scale. The project lasted from 2008-2012. Among its achievements were the monumental creation of the IMPACT dataset of historical document images with ground truth for text and layout analysis, the development of software tools for layout analysis, ground truth creation, and optical character recognition post-correction, the proposal of the PAGE format for storing document image characteristics, and the exploration of techniques for optical character recognition, layout analysis, and image correction (Papadopoulos 2013; Pletschacher & Antonacopoulos 2010; Vobl et al 2014).

The Early Modern OCR Project (eMOP)[29] was an effort by researchers at Texas A & M University to produce transcriptions of the Early English Books Online and Eighteenth

---

[25] https://diuf.unifr.ch/main/hisdoc/hisdoc2
[26] https://diuf.unifr.ch/main/hisdoc/hisdoc-iii
[27] http://www.impact-project.eu/
[28] http://www.impact-project.eu/about-the-project/concept/
[29] An overview of eMOP is available here: http://emop.tamu.edu/about

Century Collections Online databases. Containing nearly 45 million pages collectively, these two commercial databases are essential tools for historians studying the literature of the Early Modern period (fifteenth through the eighteenth century) in European history. As Christy et al. explained in their study, exemplars of significant printed works from the early modern period were microfilmed during the 1970s and 1980s, and these microfilmed images were subsequently digitized to form these databases. The initial imaging process combined with the deterioration of the original books led to bitonal images of inferior quality to modern grayscale scans. This limited the resolution and detail within the page images. The extent of the collections precludes an expensive reimaging process. Earlier attempts at commercial optical character recognition using ensemble methods had produced variable quality results. Since historical interpretation depends on accurate texts as a foundation, quality transcriptions of these significant literary works were essential. In addition to perceiving the vital need historians had for accurate transcriptions, the eMOP researchers were inspired by the aims of the IMPACT project. They hoped their work would extend and complement the results of IMPACT. Despite contributions such as the IMPACT dataset, the eMOP team believed that the aspirations of IMPACT to develop a robust optical character recognition system for historical texts had not been fully realized. In designing their own system, they therefore opted to use the open source Tesseract rather than commercial tools. They also planned to develop an historical typeface database that could form the basis for training Tesseract. The project lasted for two years and produced the following results: it produced accurate transcriptions that have been paired with the corresponding text images and made available for crowd-sourced post-correction on the 18thConnect website using the TypeWright tool, it developed a true "Big Data" infrastructure to take advantage of high performance computing resources for both optical character recognition and image post-processing, and it created additional software tools for historical document processing workflows, including Franken+ for Tesseract typeface training, Cobre for typeface identification, and Aletheia for page layout. One of their most important contributions was the pioneering work on an historical font database (Heil and Samuelson 2013). While it did not figure into the eMOP workflow as originally conceived, the researchers hope to continue to expand it and eventually have a catalog of early modern typefaces paired with the printers that utilized them as an aid to training optical character recognition systems (Christy et al 2017; Heil and Samuelson 2013).

In collaboration with the Vatican archives, the In Codice Ratio project[30] has developed a novel algorithmic segmentation technique, software tools, and a test dataset for transcribing the private correspondence of the popes. Until the inception of this digitization project, virtually none of this vast archive (estimated at 85km of linear shelving) had been scanned, much less transcribed (Firmani et al 2017). Consulting these documents, some of which date from the eighth century, required physical access by the scholar. Since these manuscripts, such as the Vatican Registers, are written in Carolingian minuscule script, the In Codice Ratio researchers at Roma Tre University approached the transcription task with a unique, hybrid methodology that drew influence from both optical character recognition and handwriting recognition. One challenge, as noted earlier in this section, for training segmentation-free handwriting recognition systems is the need for labeled data annotated by paleographers with expertise in the historical corpus language. Furthermore, the irregularities of human handwriting cause optical character recognition systems to fail. Noting these weaknesses, the In Codice Ratio researchers focused on developing a recognition system that used a segmentation lattice with the local minima of the black pixel distribution to identify sub-character "cuts" or strokes (Firmani et al 2018). A character classifier based on a convolutional neural network then identifies the cut-fragments as characters. Word prediction is accomplished with a hidden Markov model-based language model. The character classifier is trained using a custom web-interface and crowdsourcing that presents positive and negative examples of individual characters to participants who then mark unlabeled characters. Overall their system achieved an average 96% accuracy rate for individual characters and preliminary word error rates on sample pages from the Vatican collection. One of the main contributions of their work is the demonstration that traditional machine learning methods, neural networks, and a language model can be successfully incorporated into a production-grade system architecture for historical document processing.

---

[30] http://www.inf.uniroma3.it/db/icr/

## 3 Presentation of Article Classification Results

Section 4 presents the quantitative results of this literature survey. While Section 3 concentrated primarily on recent studies to provide a conceptual overview of the historical document processing field, this section offers a broader perspective to impart an empirical understanding of research in the field.

Fig. 6 offers a distribution of articles by year. It clearly indicates the growth of research in historical document processing, especially since 2006. This significant growth within the past decade corresponds with the renaissance of machine learning within computer science more generally and the extensive efforts of archival institutions to preserve their documents digitally. They also correlate with duration of the IMPACT and HisDoc family of projects from 2008-2012 and 2009-present. Furthermore, the exceptionally high article counts within the past four years from 2014-2017 also correspond with the rise of neural networks as an alternative to conventional machine learning.

Published research in the field has concentrated on OCR, layout analysis, Image quality and enhancement, binarization, and handwriting recognition. Historical document processing incorporates expertise from a variety of subfields in computer science. This diversity creates challenges for researchers, however, because synthesizing this vast quantity of material across disparate subdomains is a significant endeavor. Within the past twenty years, the majority of historical document processing research has appeared in the *IAPR International Conferences on Document Analysis and Recognition*, *International Conferences on Frontiers in Handwriting Recognition*, the *Proceedings of the IAPR International Workshops on Document Analysis Systems* and the *IAPR Workshops on Document Analysis Systems*, and the *International Conferences on Pattern Recognition*. The specialization and diversification of computer science may explain, for example, why the Ocular OCR engine has made less impact in the field than tools such as Tesseract or OCRopus. Since studies of Ocular were published in conference proceedings that focus primarily on natural language processing research, non-NLP researchers who did not follow that field remained unware of its existence or contributions to the field. The field has likewise extensively concentrated on medieval documents with less emphasis on historical documents in other periods of history. As mentioned in Section 2, while this study's focus was on Historical Document Processing primarily for documents in western languages, more research needs to be done on historical documents for languages and cultures in Africa and Asia. Expanding the historical scope of the field is just as important as enriching it as the development of new algorithms and tools. Furthermore, a preponderance of studies on computer science methods for historical document processing exist, but few are engaged by scholars in the humanities. Future research should foster greater collaboration between cultural heritage scholars and computer scientists. The results of these research endeavors should be published not just in computer science publications but in humanities journals as well.

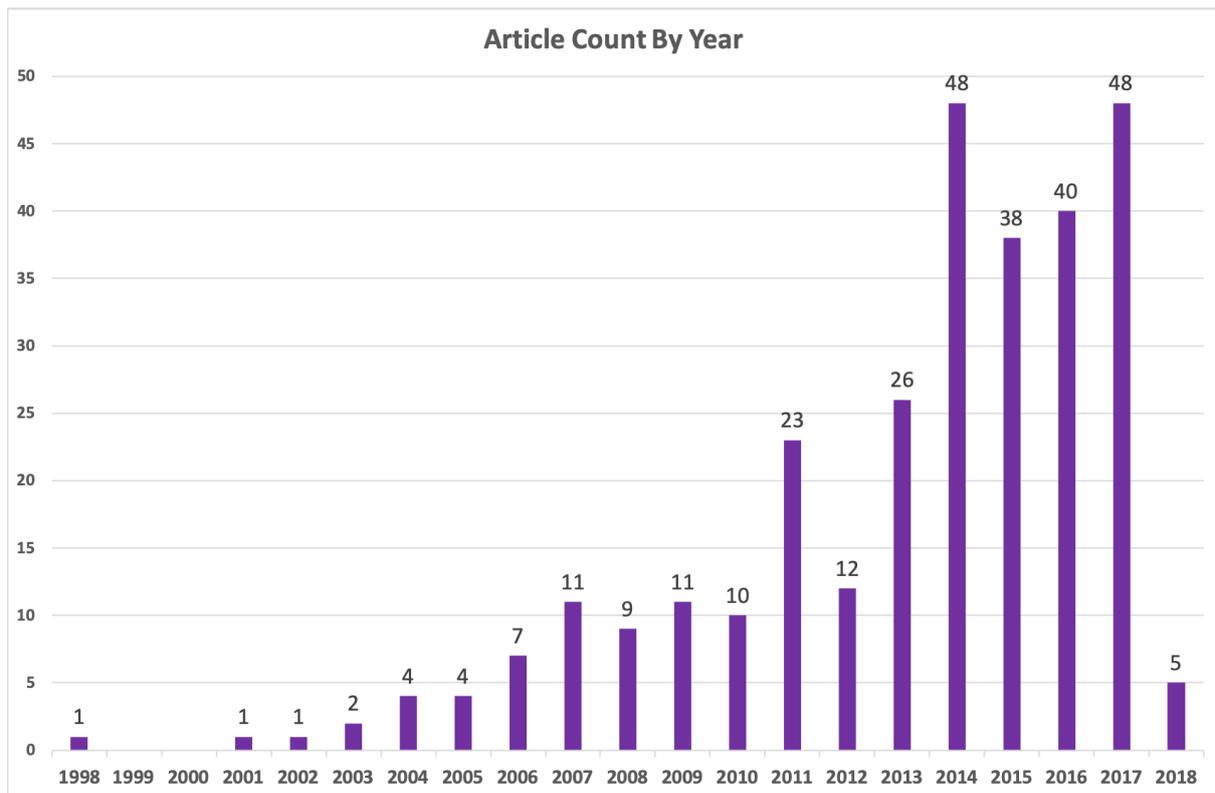

Figure 6. Article count according to year of publication

## 4 Conclusion and Directions for Further Research

Historical Document Processing transforms scanned documents from the past into digital transcriptions for the future. After document images are preprocessed through binarization, layout analysis, and line segmentation, the images of individual lines are converted into digital text through either handwritten text recognition or optical character recognition. Within the past decade, first conventional machine learning techniques using handcrafted features and more recently neural network-driven methodologies have become inherent solutions to producing accurate transcriptions from historical texts from medieval manuscripts and fifteenth century incunabula through early modern printed works. Deep learning methodologies will be an important part of creating an automated toolchain for historical document processing for eclectic and extensive historical archive collections. While they will be important for improving transcriptions, they will also be important in complex layout analysis and the creation of robust language models for historical languages. Projects such as IMPACT, Transcriptorium, eMOP, and HisDoc have made significant contributions to advancing the scholarship of the field and creating vital datasets and software tools. The combined expertise of computer scientists, digital humanists, historians, and archivists will all be necessary to meet the challenge of historical document processing for the future. As archives continue to be digitized, the volume and variety of archival data and the velocity of its creation clearly indicate that this is a "big data" challenge. Otherwise these collections will become, as the European Commission that commissioned the IMPACT project feared, "dark archives".[31] The creation of robust tools and infrastructure for this new phase of historical document processing will be the mandate of all those who wish to preserve humanity's historical textual heritage in the digital age.

---

[31] http://www.impact-project.eu/about-the-project/concept/

**5 CONCLUSION AND DIRECTIONS FOR FURTHER RESEARCH**

Historical Document Processing transforms scanned documents from the past into digital transcriptions for the future. After document images are preprocessed through binarization, layout analysis, and line segmentation, the images of individual lines are converted into digital text through either handwritten text recognition or optical character recognition. Within the past decade, first conventional machine learning techniques using handcrafted features and more recently neural network-driven methodologies have become inherent solutions to producing accurate transcriptions from historical texts from medieval manuscripts and fifteenth century incunabula through early modern printed works. Deep learning methodologies will be an important part of creating an automated toolchain for historical document processing for eclectic and extensive historical archive collections. While they will be important for improving transcriptions, they will also be important in complex layout analysis and the creation of robust language models for historical languages. Projects such as IMPACT, Transcriptorium, eMOP, and HisDoc have made significant contributions to advancing the scholarship of the field and creating vital datasets and software tools. The combined expertise of computer scientists, digital humanists, historians, and archivists will all be necessary to meet the challenge of historical document processing for the future. As archives continue to be digitized, the volume and variety of archival data and the velocity of its creation clearly indicate that this is a "big data" challenge. Otherwise these collections will become, as the European Commission that commissioned the IMPACT project feared, "dark archives". The creation of robust tools and infrastructure for this new phase of historical document processing will be the mandate of all those who wish to preserve humanity's historical textual heritage in the digital age.